# A Large Language Model Based Pipeline for Review of Systems Entity Recognition from Clinical Notes

Hieu-Tran Nghiem
*Department of Computer Science*
*Center for Health Systems Innovation*
*Oklahoma State University*
Stillwater, OK, USA
hieu.nghiem@okstate.edu

Zhuqi Miao
*Department of Management Science and Information Systems*
*Center for Heatlh Systems Innovation*
*Oklahoma State University*
Stillwater, OK, USA
zhuqi.miao@okstate.edu

Hemanth Reddy Singareddy
*Department of Computer Science*
*The State University of New York at New Paltz*
New Paltz, NY, USA
singareh2@newpaltz.edu

Jivan Lamichhane
*Department of Medicine*
*The State University of New York Upstate Medical University*
Syracuse, NY, USA
lamichhj@upstate.edu

Abdulaziz Ahmed
*Department of Health Services Administration*
*University of Alabama at Birmingham*
Birmingham, AL, USA
aahmed2@uab.edu

Johnson Thomas
*Department of Computer Science*
*Oklahoma State University*
Stillwater, OK, USA
johnson.thomas@okstate.edu

William Paiva
*Center for Health Systems Innovation*
*Oklahoma State University*
Stillwater, OK, USA
wpaiva@okstate.edu

Dursun Delen
*Department of Management Science and Information Systems*
*Center for Health Systems Innovation*
*Oklahoma State University*
Stillwate, OK, USA
dursun.delen@okstate.edu

***Abstract*—We develop a cost-effective, large language model (LLM)-based pipeline for automatically extracting Review of Systems (ROS) entities from clinical notes. The pipeline extracts ROS section from the clinical note using SecTag header terminology, followed by few-shot LLMs to identify ROS entities such as diseases or symptoms, their positive/negative status and associated body systems. We implemented the pipeline using 4 open-source LLM models: llama3.1:8b, gemma3:27b, mistral3.1:24b and gpt-oss:20b. Additionally, we introduced a novel attribution algorithm that aligns LLM-identified ROS entities with their source text, addressing non-exact and synonymous matches. The evaluation was conducted on 24 general medicine notes containing 340 annotated ROS entities. Open-source LLMs enable a local, cost-efficient pipeline while delivering promising performance. Larger models like Gemma, Mistral, and Gpt-oss demonstrate robust performance across three entity recognition tasks of the pipeline: ROS entity extraction, negation detection and body system classification (highest F1 score = 0.952). With the attribution algorithm, all models show improvements across key performance metrics, including higher F1 score and accuracy, along with lower error rate. Notably, the smaller Llama model also achieved promising results despite using only one-third the VRAM of larger models. From an application perspective, our pipeline provides a scalable, locally deployable solution to easing the ROS documentation burden. Open-source LLMs offer a practical AI option for resource-limited healthcare settings. Methodologically, our developed algorithm facilitates accuracy improvements for zero- and few-shot LLMs in named entity recognition.**

## I. Introduction

The Review of Systems (ROS) is an inventory of signs and symptoms the patient may be experiencing or has experienced [1], organized by body systems [2],[3]. It plays an important role in evidence-based assessment by guiding clinicians in prioritizing specific systems for further evaluation during the objective examination [4],[5]. Additionally, the comprehensive overview provided by ROS helps uncover underlying conditions and differentiate between diseases with overlapping symptoms, ultimately enhancing diagnostic accuracy [6]–[9].

ROS data are typically collected through verbal screenings or questionnaires at clinical encounters and are often included as a separate section within clinical notes [10]. A variety of approaches have been used to document ROS, including free-text entry, dictation, and checklists supplemented with brief free-text descriptions [3],[11]. However, documenting, reviewing, and analyzing free text within clinical notes constitute a persistent challenge to healthcare professionals. According to the literature, U.S. physicians spend 17% to 43% of their time interacting with electronic health record (EHR) systems for documentation-related tasks [12]–[14]. This significant administrative burden detracts from their primary responsibility of providing patient care, ultimately reducing productivity and diminishing career satisfaction [15]. Despite recent efforts by the American Medical Association (AMA) and the Centers for Medicare & Medicaid Services (CMS) to streamline medical documentation [16],[17], documenting medically appropriate patient history, such as history of present illness (HPI) and ROS, remain entrenched in current screening and documentation workflows, continuing to contribute to the documentation burden for health professionals [18].

Natural language processing (NLP), empowered by large language models (LLMs), offers a promising avenue for analyzing free-text clinical notes and extracting valuable information [19]–[23]. However, studies on applying NLP to ROS remains limited in the literature, and existing work has primarily relied on fine-tuned BERT models rather than LLMs [24]. BERT-based technologies typically require significant engineering effort, such as task-specific input/output processing and fine-tuning on labeled datasets, to perform well on specific tasks. In contrast, LLMs are more flexible and can be adapted to a wide range of tasks in an out-of-the-box manner using natural language prompts that are easily understood by lay users. Through in-context learning, users can apply zero-shot or few-shot instructions [25] to guide the model toward the desired behavior without additional training. Additionally, LLMs generally have significantly more parameters, enabling them to encode a broader range of knowledge within the model itself.

However, zero-/few-shot LLMs driven by natural language prompts can introduce hallucinations, rephrasing, or reordering when applied to named entity recognition (NER—in our case, ROS entity extraction). These behaviors create two challenges: (1) when extracted entities are not exact matches or do not follow the original order in the clinical notes, they require manual comparison with the gold-standard annotations; and (2) rephrased entities, even if synonymous, are still counted as errors, resulting in reduced measured accuracy.

Given the two gaps discussed above, this study proposes an easy-to-implement, productivity-enhancing pipeline that integrates open-source LLMs for ROS entity recognition on local, cost-effective devices. In addition, we developed a novel attribution algorithm that maps LLM-recognized ROS entities back to their original text locations, enabling accuracy improvements and creating the potential for future automatic annotation. Integrating this attribution algorithm into the pipeline provides a more accurate, accessible, and scalable solution for ROS entity recognition, supporting documentation efficiency for clinicians in healthcare institutions with limited resources.

## II. Related Work

### A. Clinical Entities Extraction using LLMs

Fine-tuning large language models on domain-specific corpora can enhance their performance on downstream tasks by allowing them to capture specialized vocabulary and contextual nuances. Earlier research in NLP primarily focused on fine-tuning BERT-based models for downstream tasks. Popular models like BioBERT [26], ClinicalBERT [27] or recently GatorTron [28] have demonstrated state-of-the-art performance on Named Entity Recognition (NER) tasks across biomedical and clinical domains. With the recent growth of open-source, generative LLMs models such as Llama [29], fine-tuning on these models can yield even better performance, but with the cost of substantially greater computational requirements [30].

Given the computing power and data annotation demands of model fine-tuning, we adopt a more resource-efficient learning strategy. Zero-shot and few-shots learning settings in LLMs have been increasingly adopted in clinical NLP tasks such as entity recognition and information extraction [31],[32]. These tasks are performed through in-context learning, using only instructions or examples provided within the prompt, without any weight updates or training. The model relies purely on its general knowledge and understanding of language patterns from pretraining, so it can perform tasks with little or no explicit training data, making it well suited for limited resources environments.

### B. Atrribution

Attribution refers to the ability of the model, such as an LLM, to accurately use documents, or generate and provide evidence, often in the form of references or citations, to support its generated text [33]. Attribution increases transparency by making the model output comprehensible and verifiable to general audience, while also helping reduce issues such as hallucination and reasoning errors. Attribution approaches often depend on further fine-tuning or rely on external data sources to enhance a model's ability to ground its output. Yue et al. (2023) [34] improve attribution through fine-tuning on high-quality datasets, while Huang et al. (2024) [35] introduce the START framework, which refines model responses via self-improvement and Rejection Sampling Fine-Tuning, with evaluating candidate outputs based on attributability, robustness, and comprehensiveness. Patel et al. (2024) [36] builds a multi-source attribution dataset (MULTIATTR) by transforming existing QA datasets so that model outputs can be linked to multiple sources for factual grounding.

In contrast, for medical entity extraction, the original clinical note itself provides the most authoritative and context-accurate source of truth. Because every extracted entity must correspond to language explicitly present in the note, attribution can be performed directly against the source document without requiring additional databases or external evidence. This approach ensures that each model-identified span is grounded in the clinician's original wording, preserves the integrity of the clinical narrative, and avoids the noise or mismatch that can arise when relying on outside data sources.

## III. Methods

### A. Clinical Notes

We employed Medical Transcription Sample Reports and Examples (MTSamples)[1] as our data source. MTSamples is a publicly available repository of de-identified clinical notes that is widely used in the medical informatics research community [37]. The notes in MTSamples are organized into sections, each with a header followed by free text, as illustrated in Fig. 1(A).

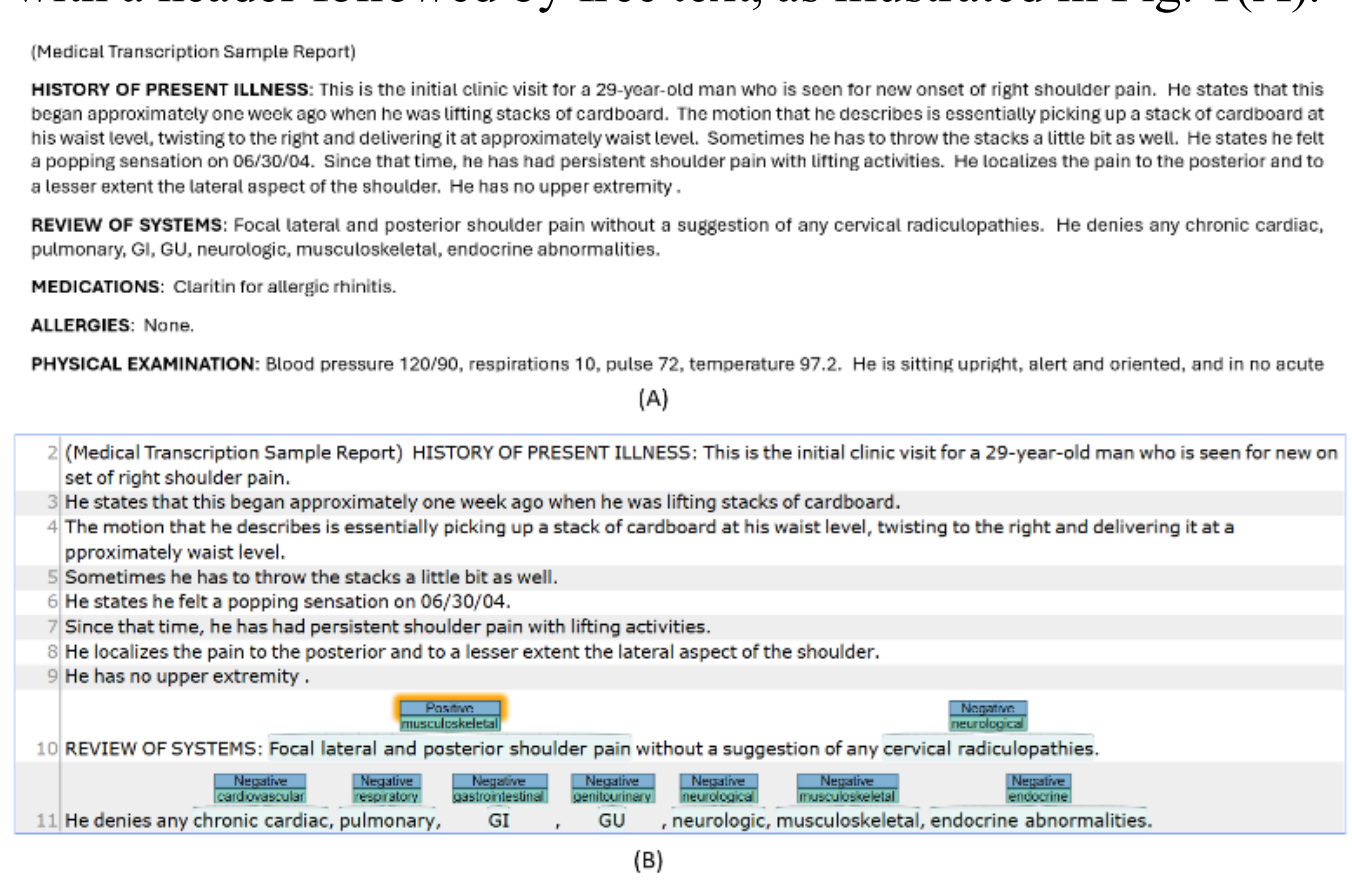

Fig. 1. An example of MTSamples notes: (A) The sample note in plain text format; (B) The sample note with annotations.

[1] https://mtsamples.com

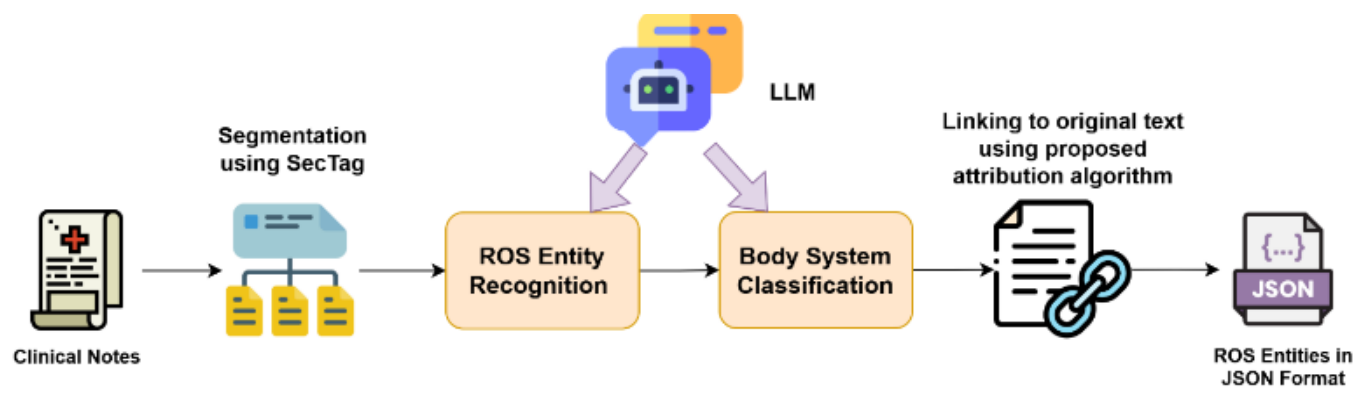

Fig. 2. Overview of the proposed pipeline

We collected and annotated a high-quality dataset for our experiment. First, we selected 24 general medicine notes from MTSamples that included a ROS section. The sample comprised of 15 consultation notes, 4 ED notes, 3 progress notes and 2 discharge summaries. Of these, 12 notes contained simplistic ROS sections lacking details, using phrases such as "Noncontributory," "Otherwise negative," or "As per the HPI." Including these types of notes enhances the dataset's diversity and allows for a better evaluation of our pipeline in handling varied styles in the ROS entity recognition task.

From the selected notes, we annotated n=340 entities in the ROS sections, including diseases, symptoms, their negation statuses, and body systems. The annotations were first completed by two undergraduate biology major students in a double-blind manner, and then curated by two professors (one of them is the co-author Z.M), followed by verification and adjustment by an experienced doctor. Each disease or symptom entity was labeled with problem, then its status (positive or negative), and finally assigned to one of the 14 standard body systems: constitutional, eyes, ENT (ears, nose, mouth, and throat), cardiovascular, respiratory, hematologic/lymphatic, gastrointestinal, genitourinary, musculoskeletal, integumentary, neurological, psychiatric, endocrine, and allergic/immunologic [38]. The annotation process resulted in a total of 340 spans and 1020 labels, as exemplified in Fig. 1(B).

### B. *Pipeline Design*

Our pipeline, illustrated in Fig. 2, comprises four consecutive steps: Segmentation, ROS Entity Recognition, Body System Classification and final attribution. Each of the four steps is further described below:

- Segmentation: This step segments clinical notes and extracts the ROS section to ensure that downstream steps focus specifically on this section. Specifically, we used the SecTag section header terminology [39],[40] to identify the start and end boundaries of the ROS section and extract the corresponding text.
- ROS Entity Recognition: This step employs few-shot LLMs [25],[32] to extract ROS entities such as diseases or symptoms from the isolated ROS section. The LLMs are then instructed to perform negation detection or evaluate the present or absent status of each entity (i.e., positive or negative, in medical perspective) based on the context. The few-shot examples were provided not only to guide the recognition process but also to illustrate the desired JSON output format.
- Body System Classification: This step classifies the extracted ROS entities by body system. We employed few-shot LLMs to implement the classification. Since many diseases and symptoms are self-indicative of their corresponding systems, the few-shot examples primarily serve to guide the model toward generating consistent disease–system pairs, a format easy to parse using regular expressions [41]. The resulting systems are then added to the JSON outputs from the previous step. Note that we separated this classification step from the prior Disease and Symptom Recognition step to avoid overly lengthy prompts, which in our preliminary experiments increased confusion and reduced model accuracy.
- Attribution algorithm: Once the body systems are identified, this step performs a revalidation process using our proposed attribution algorithm to correct the LLM's output and map all rephrased or reordered entities back to the original text sequence. We will elaborate its mechanism in the next section.

### C. *Attribution Algorithm*

We propose an attribution algorithm which employs efficient embedding technique and semantic similarity scoring to link model outputs back to the original entities, relying on the original clinical notes as the ground-truth source to ensure maximal factual reliability. Without referring to third-party datasets, manually curated knowledge bases or external search tools, it makes the system more robust, private and easier to maintain.

The core idea of our proposed attribution algorithm is to locate the *n-gram* within a document span that is most similar to the LLM-generated ROS entity. An *n-gram* is a contiguous sequence of *n* items, usually words taken from a text [42]. As illustrated in Fig. 3, given a document represented as a sequence of *m* words, and an LLM-recognized entity that does not exactly match any contiguous token sequence, the algorithm enumerates all possible *n-grams* $(n = 1, 2, \ldots, m)$ and computes an embedding for each. Next, it measures the embedding-based cosine similarity between every *n-gram* and the LLM-produced entity. The *n-gram* with the highest similarity score is then selected as the best candidate for the original text span to which the entity should be attributed.

### D. *Implementation and Evaluation*

To develop a simple and cost-effective pipeline, we adopted medium- or small-sized open-source, quantized LLMs that can run on consumer/workstation-grade GPUs to perform ROS entity extraction. Specifically, we employed mistral-small 3.1 [43] (24B parameters, size of 15GB, Q4_K_M quantization), llama 3.1 [44] (8B, 5GB, Q4_K_M), gemma3 [45] (27B, 17GB, Q4_K_M), and gpt-oss-20b [46] (20B, 14GB, MXFP4). We integrated these models into our workflow using Ollama [47] together with LangChain [48], which offers a convenient interface for constructing LLM pipelines. For the LLM few-shot configuration, we set temperature = 0 to enforce determinism, with seed = 42, top-k = 10, and top-p = 0.5. Detailed prompts used in the proposed pipeline are provided in Supplementary Material 1.

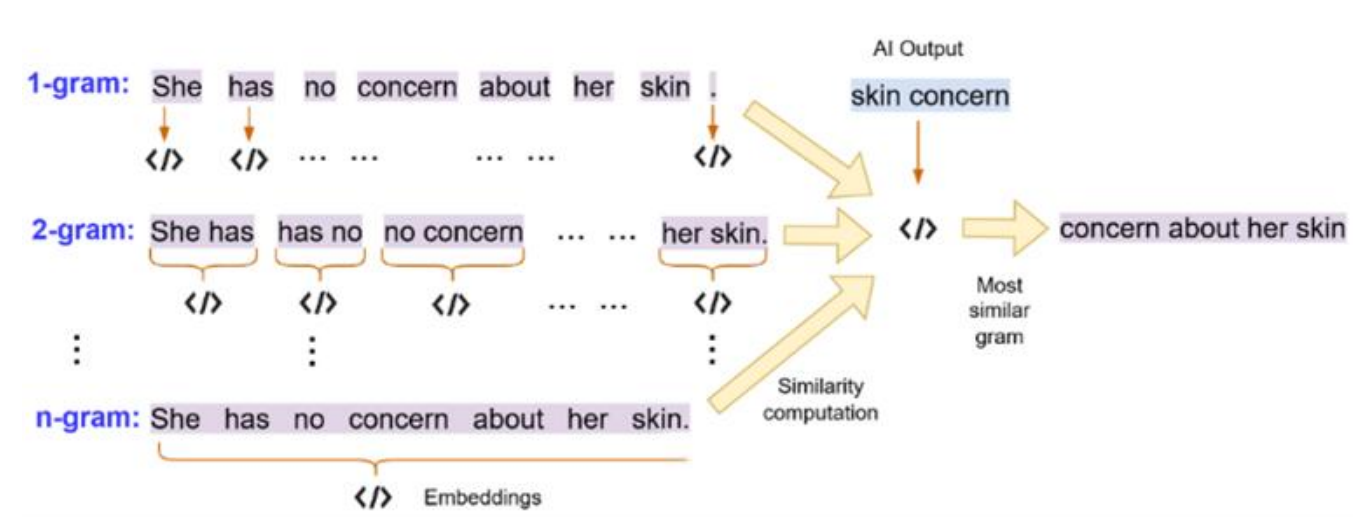

Fig. 3. Illustration of the proposed attribution algorithm. The original ROS-related text is "concern about her skin", while the LLM identified the same concept but rephrased it as "skin concern." The algorithm evaluates all n-grams in the source text and identifies that "concern about her skin" has the highest similarity, thereby recovering the correct span as the exact match.

In our implementation of the attribution algorithm, we used the all-MiniLM-L6-v2[2] sentence-transformers model which brings excellent balance of speed, accuracy and efficiency to compute embeddings for each *n-gram* and the LLM outputs. The text is mapped into a 384-dimensional vector space which is suitable for semantic comparison, and cosine similarity scores are calculated to identify the *n-gram* with the highest similarity score. All LLMs inference, embedding computation, and similarity calculations were performed using Python 3.12 and sentence-transformer 5.1.2 on a consumer-grade NVIDIA RTX 3090 GPU with 24GB of VRAM. Our source code for the pipeline, along with annotated notes, is available on GitHub (see the Data and Code Availability Statement).

In the proposed pipeline, which evaluates multiple LLMs for recognizing ROS entities, performance is assessed both with and without the attribution algorithm across three tasks: ROS entity extraction, negation detection, and body-system classification. The attribution algorithm helps correct entity errors arising from the model's rephrasing or reordering of the original text. We report standard evaluation metrics for entity extraction – under both exact-match and relax-match criteria: precision, sensitivity, and F1 score - following the conventional evaluation framework established in foundational NLP benchmarks such as the CoNLL shared tasks (Sang et al. 2003) [49]. Exact match (true positive) requires that the predicted entity span and its entity types align perfectly to the gold-standard label, whereas relaxed match requires only entity type matches, with overlapping entity's text span. We also evaluated two error types: over detections (false positive), where the model labels unannotated text as an ROS entity, or the model output rephrased or reordered text compared to original entity span sequence; under detections (false negative), where annotated ROS entity is missed. Accuracy - defined as the proportion of exact or relaxed matches out of all annotated entities, and error rate - reflecting the number of false positives and false negatives relative to all annotated entities - together indicate the extent of manual correction required after applying the tool and serves as an important performance measure in clinical perspective.

## IV. Results

In general, two largest models - gemma3:27b and mistral3.1:24b - outperformed other models in terms of standard NER benchmark F1 scores as shown in Table I. Notably, every model benefited from the attribution algorithm, which enhanced performance across all three tasks in the pipeline as illustrated in Fig. 4 . Detailed results on precision, sensitivity and F1 score of each task for all models are provided in Supplementary Table 3.

Table II summarizes the accuracy and error rate of the pipeline based on each LLM; in addition, the performance with and without our attribution algorithm have been also compared and shown in the table. In the first NER task of the pipeline - ROS entity recognition, when we ran without attribution, 67.4% to 81.8% of ROS entities are exact-matched (with respective to the total number of annotations, varying on LLM used), and 77.4% to 93.2% are relax-matched. The error rate ranges from 23.9% to 39.4% if we require exact matches strictly, and from 16.3% to 30.4% when we accept relaxed matches. With attribution integrated, the exact match rates were enhanced to 75.9-85.6%, while the relaxed match rates increased to 88.8-97.6%. Meanwhile, the error rates dropped to 17.1-26.1% and 9.1-14.0%, for exact and relaxed matches respectively.

Once the ROS entities are identified, the LLMs continue to demonstrate strong performance in detecting their positive / negative status and classifying them into proper body systems. Fig. 5 shows that all models (except the smaller model llama3.1) achieved over 90% accuracy in detecting negation status for relax-matched entities, with gpt-oss reaching 96.5%. For body system classification, LLMs also performed well despite slight drop, with accuracy ranging from 76.5% to 83.8%. In general, consistent with the results from the first task, the attribution-enhanced method achieved higher accuracy and lower error rates than the non-attribution approach in the second and third tasks - negation detection and body system classification. For all tasks, the error rates of all models with attribution are statistically significant compared to without attribution on the one-sided Wilcoxon rank-sum / Mann–Whitney U test at a significance level of α = 0.05. Detailed results are provided in Supplementary Table 3.

TABLE I. F1 Score In Three Tasks Of The Proposed Pipeline: ROS Entity Recognition (ROS), Negation Detection (NEG) And Body System Classification (BD). Bold Scores Indicate Performance Improvement With Attribution Compared To Without Attribution. Underlined Score Is The Best F1 Score Among 4 Models

| Model | EXACT MATCH | | | | | |
|---|---|---|---|---|---|---|
| | ***With attribution*** | | | ***Without attribution*** | | |
| | ROS | NEG | BD | ROS | NEG | BD |
| llama3.1:8b | **0.850** | **0.833** | **0.788** | 0.755 | 0.748 | 0.701 |
| gemma3:27b | **0.906** | **0.894** | **0.856** | 0.864 | 0.860 | 0.818 |
| mistral3.1:24b | **0.884** | **0.877** | **0.797** | 0.827 | 0.825 | 0.749 |
| gpt-oss:20b | **0.860** | **0.855** | **0.777** | 0.820 | 0.817 | 0.736 |
| | **RELAX MATCH** | | | | | |
| llama3.1:8b | **0.928** | **0.913** | **0.854** | 0.821 | 0.813 | 0.763 |
| gemma3:27b | **0.951** | **0.939** | **0.898** | 0.912 | 0.905 | 0.860 |
| mistral3.1:24b | **0.952** | **0.944** | **0.854** | 0.894 | 0.889 | 0.800 |
| gpt-oss:20b | **0.925** | **0.919** | **0.832** | 0.884 | 0.878 | 0.790 |

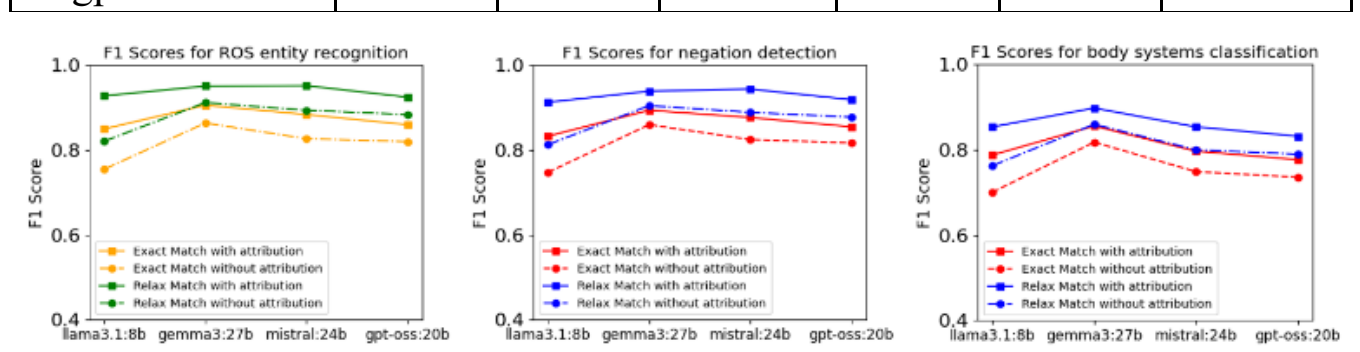

Fig. 4. Performance comparison between models: with attribution and without attribution

2. https://www.sbert.net/docs/sentence_transformer/pretrained_models.html

TABLE II. ROS ENTITY RECOGNITION PERFORMANCE OF THE PIPELINE BY LLMS: COUNT OF MATCHES AND ERRORS; ACCURACY AND ERROR RATE BETWEEN MODELS - WITH ATTRIBUTION AND WITHOUT ATTRIBUTION. FOR OTHER TWO TASKS (NEGATION DETECTION AND BODY SYSTEM CLASSIFICATION), PLEASE REFER TO SUPPLEMENTARY TABLE 3

| **Without Attribution** | | **Models** | | | |
|---|---|---|---|---|---|
| ***Matching Criteria*** | ***Evaluation metric[a]*** | ***Llama3.1 :8b*** | ***Gemma3 :27b*** | ***Mistral3. 1:24b*** | ***Gpt-oss:20b*** |
| Exact Match | EM/ Accuracy | 229 (67.4%) | 276 (81.2%) | 260 (76.5%) | 278 (81.8%) |
| | OD | 69 | 44 | 62 | 95 |
| | UD | 80 | 43 | 47 | 27 |
| | ER | 39.4% | 23.9% | 29.5% | 30.5% |
| Relax Match | RM/ Accuracy | 263 (77.4%) | 304 (89.4%) | 298 (87.6%) | 317 (93.2%) |
| | OD | 35 | 16 | 24 | 56 |
| | UD | 80 | 43 | 47 | 27 |
| | ER | 30.4% | 16.3% | 19.2% | 20.8% |
| **With Attribution** | | | | | |
| Exact Match | EM/ Accuracy | 258 (75.9%) | 290 (85.3%) | 277 (81.5%) | 291 (85.6%) |
| | OD | 48 | 32 | 47 | 84 |
| | UD | 43 | 28 | 26 | 11 |
| | ER | 26.1% | 17.1% | 20.9% | 24.6% |
| Relax Match | RM/ Accuracy | 302 (88.8%) | 317 (93.2%) | 318 (93.5%) | 332 (97.6%) |
| | OD | 4 | 5 | 6 | 43 |
| | UD | 43 | 28 | 26 | 11 |
| | ER | 13.5% | 9.4% | 9.1% | 14.0% |

[a] EM: exact matches, RM: relax matches, OD: over detections, UD: under detections, ER: error rate

In addition, we observed several common patterns as follows:

- Many relaxed matches resulted from entity rephrasing. For example, "fevers" was detected as "fever," and "concerns about her skin" was matched as "skin concerns". They have been fixed by our attribution algorithm.
- Some output entities were entirely hallucinated. For example, given the text "REVIEW OF SYSTEMS: As above. No nausea, vomiting, or diarrhea.", the LLMs detected "fever", "headache" and "back pain". Our algorithm mapped these entities back to the most similar text spans: "nausea" and "vomiting".
- When a sequence of words consists of multiple entities, LLMs sometimes detect only one, or reorder the entity span. For example, with "darkening of the skin or eyes", LLMs only detected "darkening of the eyes". Our algorithm corrected the original output to "darkening of the skin or eyes" resulting in an extra match.
- One clinical note had a ROS section that was only partially captured by SecTag, resulting in 15 under-detected entities.

Furthermore, it is worth noting that llama model utilizes only 5GB of GPU memory, in contrast to gpt-oss's 14GB, mistral's 15GB and gemma's 17GB. As a result, llama's inference time is significantly faster than the other three open-source LLMs. Subsequently, the attribution algorithm has an average execution time of approximately 2 seconds per note, as shown in Table III. Both embedding calculation and similarity

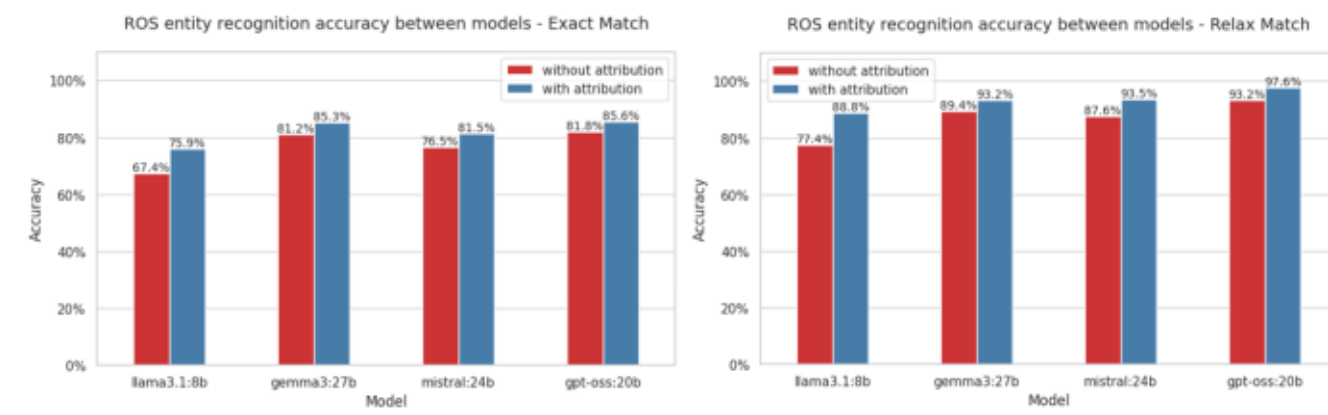

*(A) ROS entity recognition accuracy with attribution versus no attribution among 4 LLMs*

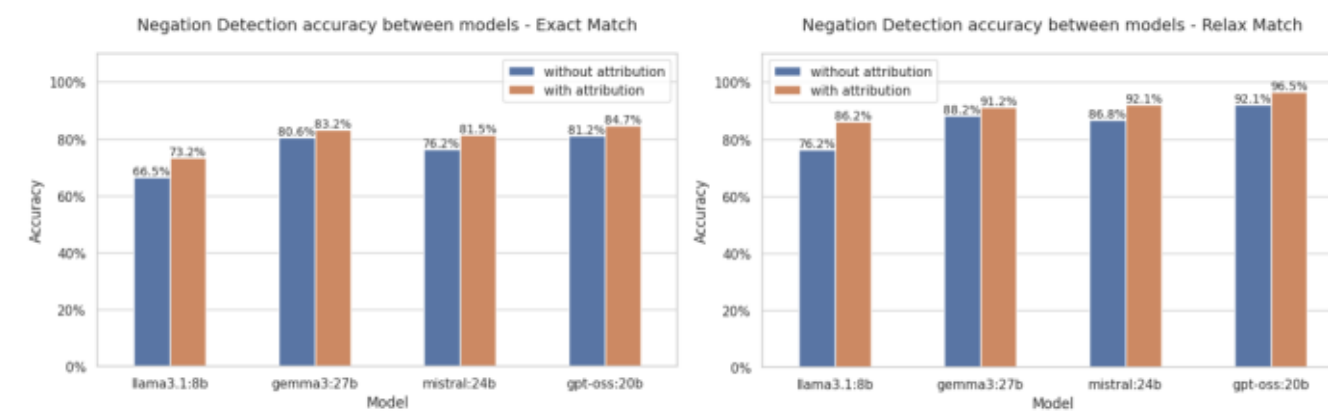

*(B) Negation detection accuracy with attribution versus no attribution among 4 LLMs*

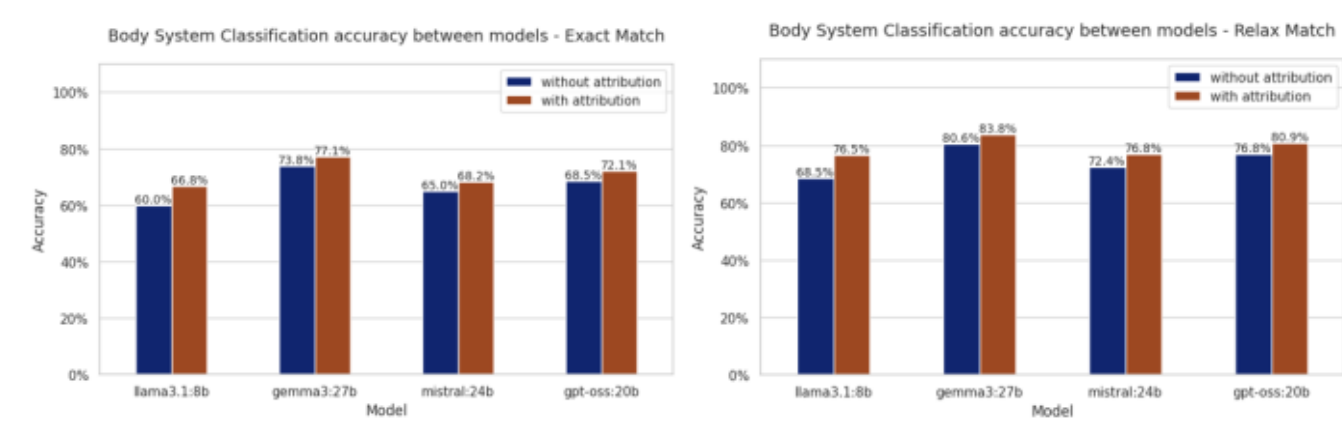

*(C) Body system classification accuracy with attribution versus no attribution among 4 LLMs*

Fig. 5. The accuracy of ROS entity recognition, negation detection and body system classification of the pipeline with different LLMs

TABLE III. EXECUTION TIME OF ATTRIBUTION ALGORITHM FOR VALIDATING EXTRACTED ENTITIES TO ORIGINAL TEXT SEQUENCE. THE ROS SECTIONS ACROSS 24 NOTES CONTAIN TOTAL OF 1560 WORDS

| **Model** | **Number of entities extracted** | **Attribution time (seconds)** | **Attribution time per note (seconds)** |
|---|---|---|---|
| llama3.1:8b | 291 | 52.72 | 2.20 |
| gemma3:27b | 320 | 53.35 | 2.22 |
| mistral3.1:24b | 316 | 54.06 | 2.25 |
| gpt-oss:24b | 365 | 54.03 | 2.25 |

computation are executed on one consumer-grade GPU, enabling efficient parallel computation for optimal runtime. Efficiency is especially valuable when computational resources are limited in pipeline deployment.

## V. DISCUSSION

Documentation has long been a burden for health professionals. Although the AMA and CMS have simplified requirements, specifically by removing the need to document detailed history and examination, clinical documentation still consumes a significant amount of time in practice. Much of this time is spent extracting key information to complete claim templates.

This study presents a LLM-based pipeline in combination with a novel attribution algorithm that is capable of automatically batch-processing clinical notes to extract ROS entities, determine their status, classify them into corresponding body systems and more importantly, link the output entities to the evidence in the original text. This process is essential for entity extraction with LLMs, as preserving the original text sequence is critical. The pipeline can be used with moderately sized, open-source LLMs on lightweight devices that can be easily deployed within the local infrastructure of healthcare institutions, thereby protecting patient data and offering a cost-effective solution. Our evaluation demonstrated the pipeline's generalizability and efficiency. When integrated into the pipeline, all evaluated LLMs succeeded in reducing the ROS entity recognition workload remarkably compared to a fully manual process: up to 85.6% and 97.6% ROS entities can be detected exactly or relaxed respectively. When detected, up to 84.7% of exact matched entities can be identified correctly for positive/negative status and up to 77.1% be classified into the correct body systems. For relaxed matched entities, the measures are 96.5% and 83.8%.

Our results also showed the variability in performance that may arise from the choice of underlying LLMs. Key considerations when selecting an LLM include accuracy, execution efficiency, device requirements, and usage cost. Llama is a lightweight yet effective option that is well-suited for more resource-constrained environments. When resources permit, other bigger models will present a stronger alternative. Similar to the burden of recognizing and documenting ROS, documenting HPI and Physical Examinations poses an even greater challenge for health professionals [18]. Our current work on ROS extraction can be extended to identify HPI-associated signs and symptoms, helping to streamline the HPI documentation process. Another interesting technical enhancement for future work is to improve the efficiency of our attribution algorithm by reducing its execution time. Complementary to this, mitigating the LLM's rephrasing tendency would enable more precise matching, thereby further improving overall performance.

Limitations: The clinical notes used in this study were from a single data source and focused on general medicine. Note formats and content can vary across institutions and specialties. Evaluating the pipeline on a broader range of samples will be instrumental in enhancing its generalizability and performance. The pipeline's performance is primarily measured by error rates, which help estimate how many ROS entities may require manual correction. However, this is only an approximation. The actual time savings achievable in real-world settings should be evaluated through practical trials.

## VI. Conclusion

This study is the first to directly tackle the challenge of ROS entity extraction - a process that often requires considerable clinician time. An open-source LLM-based pipeline for automating ROS entity recognition with a validation component is developed and evaluated in this study. The pipeline is cost-effective, straightforward to implement and can integrate various open-source LLMs to achieve comparable accuracy as state-of-the-art models. Future work will be focused on accuracy improvement and adaptation to HPI (History of Present Illness) entity recognition.

## Acknowledgment

We thank Ellie P. Nguyen, Suhao Chen, and Andrew Gin for their valuable contributions to the annotation process.

## Data And Code Availability

Original datasets can be found at https://mtsamples.com. High-resolution versions of the figures are provided in the Supplementary Material. All annotated notes, source code and Supplementary Material supporting this study have been made publicly available at:

https://github.com/hieutrann/ROS_Entity_Extraction_Pipeline